\documentclass[sigconf]{acmart}
\AtBeginDocument{%
  \providecommand\BibTeX{{%
    \normalfont B\kern-0.5em{\scshape i\kern-0.25em b}\kern-0.8em\TeX}}}

\copyrightyear{2023}
\acmYear{2023}
\setcopyright{acmlicensed}\acmConference[ICMR '23]{International Conference on Multimedia Retrieval}{June 12--15, 2023}{Thessaloniki, Greece}
\acmBooktitle{International Conference on Multimedia Retrieval (ICMR '23), June 12--15, 2023, Thessaloniki, Greece}
\acmPrice{15.00}
\acmDOI{10.1145/3591106.3592278}
\acmISBN{979-8-4007-0178-8/23/06}

\begin{document}

\title{Framing the News: \\ From Human Perception to Large Language Model Inferences }


\author{David Alonso del Barrio}
\email{ddbarrio@idiap.ch}
\affiliation{%
  \institution{Idiap Research Institute}
  \country{Switzerland}
}

\author{Daniel Gatica-Perez}
\email{gatica@idiap.ch}
\affiliation{%
  \institution{Idiap Research Institute and EPFL}
  \country{Switzerland}
}


\begin{abstract}
\footnote{\textbf{David Alonso del Barrio, Daniel Gatica-Perez| ACM 2023. This is the author's version of the work. It is posted here for your personal use. Not for redistribution. The definitive Version of Record will be published in https://doi.org/10.1145/3591106.3592278.}}
Identifying the frames of news is important to understand the articles' vision, intention, message to be conveyed, and which aspects of the news are emphasized. Framing is a widely studied concept in journalism, and has emerged as a new topic in computing, with the potential to automate processes and facilitate the work of journalism professionals. In this paper, we study this issue with articles related to the Covid-19 anti-vaccine movement. First, to understand the perspectives used to treat this theme, we developed a protocol for human labeling of frames for 1786 headlines of No-Vax movement articles of European newspapers from 5 countries. Headlines are key units in the written press, and worth of analysis as many people only read headlines (or use them to guide their decision for further reading.)
Second, 
considering advances in  Natural Language Processing (NLP) with large language models, we investigated two approaches for frame inference of news headlines: first with a GPT-3.5 fine-tuning approach, and second with GPT-3.5 prompt-engineering.  Our work contributes to the study and  analysis of the performance that these models have to facilitate journalistic tasks like classification of frames, while understanding whether the models are able to replicate human perception in the identification of these frames.
\end{abstract}






\maketitle

\section{Introduction}

In recent years, there has been a proliferation in the use of concepts such as data journalism, computational journalism, and computer-assisted reporting \cite{coddington2015clarifying} \cite{hermida2017finding}, which all share the vision of bridging journalism and technology. 
The progress made in NLP 
has been gradually integrated into the journalistic field \cite{biswal2020artificial}\cite{broussard2019artificial}\cite{articlecscw}.
More specifically, machine learning models based on transformers have been integrated in the media sector in different tasks \cite{milosavljevic2021our} such as the creation of headlines with generative languages models \cite{dale2021gpt}, summarization of news articles \cite{gupta2022automated}\cite{grail2021globalizing}, false news detection \cite{RAI202298}, and 
topic modeling and sentiment analysis \cite{ghasiya2021investigating}.
The development of large language models such as GPT-3 \cite{brown2020language}, BLOOM \cite{scao2022bloom} or ChatGPT show a clear trend towards 
human-machine interaction becoming easier and more intuitive, opening up a wide range of research possibilities. At the same time, the use of these models is also associated with a lack of transparency 
regarding how these models work, but  efforts are being made to bring some transparency to these models, and to analyze use cases where they can be useful and where they cannot \cite{holistic}. 
Based on the premises that these models open up a wide range of research directions \cite{gpt_bar_exam}, and that at the same time (and needless to say) they are not the solution to all problems, we are interested in 
identifying use cases and tasks where they can be potentially useful, while acknowledging and systematically documenting 
their limitations \cite{tamkin2021understanding}. 
More specifically, the aim of this work is to analyze the performance of GPT-3.5 for a specific use case, namely the analysis of frames in news, from an empirical point of view, with the objective of shedding light on a potential use of generative models in journalistic tasks.

Frame analysis is a concept from journalism,
which consists of studying the way in which news stories are presented on an issue, and what aspects are emphasized: Is a merely informative vision given in an article? Or is it intended to leave a moral lesson? Is a news article being presented from an economic point of view? Or from a more human, emotional angle? The examples above correspond to different frames with which an article can be written.

The concept of news framing has been studied in computing as a step beyond topic modeling and sentiment analysis, and for this purpose, in recent years, pre-trained language models have been used for fine-tuning the classification process of these frames \cite{yla2022topic} \cite{burscher2014teaching}, but the emergence of generative models opens the possibility of doing prompt-engineering of these classification tasks, instead of the fine-tuning approach investigated so far.


Our work aims to address this research gap by posing the following research questions:

{\bf RQ1}: What are the main frames in the news headlines about the anti-vaccine movement, as reported in newspapers across 5 European countries?

{\bf RQ2}: Can prompt engineering be used for classification of headlines according to frames? 



By addressing the above research questions, our work makes the following contributions:

{\bf Contribution 1.} We implemented a process to do human annotation of the main frame of 1786 headlines of articles about the Covid-19 no-vax movement, as reported in 19 newspapers from 5 European countries (France, Italy, Spain, Switzerland and United Kingdom.) At the headline level, we found that the predominant frame was human interest, where this frame corresponds to a personification of an event, either through a statement by a person, or the explanation of a specific event that happened to a person. Furthermore, we found 
a large number of headlines annotated as containing no frame, as they simply present information without entering into evaluations. 
We also found that for all the countries involved, the distribution of frame types was very similar, i.e.,  
human interest and no frame are the two predominant frames. Finally, the generated annotations allowed to subsequently study the performance of a large language model. 

{\bf Contribution 2.} 
We studied the performance of GPT-3.5 on the task of frame classification of headlines. In addition to using the fine-tuning approach from previous literature,  we propose an alternative approach for frame classification that requires no labeled data for training, namely prompt-engineering using GPT-3.5. The results show that fine-tuning with GPT-3.5
produces 72\% accuracy (slightly higher than other smaller models), and that the prompt-engineering approach results in lower performance (49\% accuracy.) Our analysis also shows that the subjectivity of the human labeling task has an effect on the obtained accufracy.


The paper is organized as follows. In Section \ref{section:related_work}, we discuss related work. In Section \ref{section:data}, we describe the news dataset. In Section \ref{section:methodology}, we describe the methodology for both human labeling and machine classification of news frames. We present and discuss results for RQ1 and RQ2 in Sections \ref{section:results_discussion} and \ref{section:results_discussion2}, respectively.
Finally, we provide  conclusions in Section \ref{section:conclusions}.

\section{Related Work} \label{section:related_work}
Framing has been a concept widely studied in journalism, with a definition that is rooted in the study of this domain \cite{entman1993framing}: “To frame is to select some aspects of a perceived reality and make them more salient in a communicating text, in such a way as to promote a particular problem definition, causal interpretation, moral evaluation, and/or treatment recommendation for the item described.” 


For frame recognition, there are two main approaches: the inductive approach \cite{cooper2010oppositional}, where one can extract the frames after reading the article, and the deductive approach \cite{deductive},  where a predefined list of frames exists and the goal is to interpret if any of them appears in the article. In the deductive case, there are generic frames and subject-specific frames, and the way to detect them typically involves reading and identifying one frame at a time, or through answers to yes/no questions that represent the frames. Semetko et al. \cite{semetko} used 5 types of generic frames (attribution of responsibility, human interest, conflict, morality, and economic consequences) based on previous literature, and they defined a list of 20 yes/no questions to detect frames in articles. For instance, the questions about morality are the following: "Does the story contain any moral message?
Does the story make reference to morality,
God, and other religious tenets?
Does the story offer specific social
prescriptions about how to behave?", and so on for each of the frame types.
This categorization of frames has been used in various topics such as climate change \cite{climate_change_frame} \cite{dirikx2010frame}, vaccine hesitance \cite{catalan2020vaccine}, or immigration \cite{kim2018news}. 

We now compare the two approaches on a common topic, such as Covid-19.
Ebrahim et al. \cite{ebrahim2022corona} 
followed an inductive approach in which the frames were not predefined but emerged from the text (e.g., deadly spread, stay home, what if, the cost of Covid-19) using headlines as the unit of analysis.
In contrast, the deductive approach has studied very different labels. El-Behary et al. \cite{el2021feverish} followed the method of yes/no questions, but in addition to the 5 generic frames presented before, they also used blame frame and fear frame. 
Adiprasetio et al. \cite{adiprasetio2020pandemic} and Rodelo  \cite{rodelo2021framing} used the 5 generic frames with yes/no questions, while Catalán-Matamoros et al. \cite{catalan2019media} used the 5 frames and read the headline and subheadline to decide the main frame. Table 1 summarizes some of the the existing approaches.
This previous work showed how frame labels can be different, and also that frame analysis has been done at both headline and article levels.
These two approaches (inductive and deductive) that originated in journalism have since been replicated in the computing literature.

We decided to follow the deductive approach because a predefined list of frames allows to compare among topics, countries, previous literature, and also because
they represent a fixed list of labels for machine classification models.
Furthermore, the inductive approach tends to be more specific to a topic, and from the computing viewpoint, past work has tried to justify topic modeling as a technique to extract frames from articles.

Ylä-Antitila et al. \cite{yla2022topic} proposed topic modeling as a frame extraction technique. They argued that topics can be interpreted as frames if three requirements are met: frames are operationalized as connections between concepts; subject-specific data is selected; and topics are adequately validated as frames, for which they suggested a practical procedure. This approach was based on the choice of a specific topic (e.g., climate change) and the use of Latent Dirichlet Allocation (LDA) as a technique to extract a number of subtopics. In a second phase, a qualitative study of the top 10 words of each subtopic was performed, and the different subtopics were eliminated or grouped, reducing the number and establishing a tentative description. In a third phase, the top 10 articles belonging to that frame/topic were taken, and if the description of the topic fitted at least 8 of the 10 articles, that topic/frame remained. The frames found in this article were: green growth, emission cuts, negotiations and treaties, environmental risk, cost of carbon emissions, Chinese emissions, economics of energy production, climate change, environmental activism, North-South burden sharing, state leaders negotiating, and citizen participation.

From Entman's definition of frame \cite{entman1993framing}, it seems that the deductive approach is more refined than the inductive approach (which seems to resemble the detection of sub-themes.) For example, with regard to climate change, there are stories on how people have been affected by climate change from an emotional point of view, thus personalizing the problem. In this case,  we could categorize the corresponding frame as human interest, as the writer of the article is selecting "some aspects of a perceived reality and make them more salient". The language subtleties with which news articles are presented cannot be captured with basic topic modeling.

Isoaho et al.\cite{isoaho2021topic} held the position that while the benefits of scale and scope in topic modeling were clear, there were also a number of problems, namely that topic outputs do not correspond to the methodological definition of frames, and thus topic modeling remained an incomplete method for frame analysis. Topic modeling, in the practice of journalistic research, is a useful technique 
to deal with the large datasets that are available, yet is often not enough to do more thorough analyses  \cite{jacobi2016quantitative}.
In our work, we clearly notice that frame analysis is not topic modeling. For example, two documents could be about the same topic, say Covid-19 vaccination, but one article could emphasize the number of deaths after vaccination, while the other emphasized the role of the vaccine as a solution to the epidemic. 

We also consider that the larger the number 
of possible frame types, the more likely it is to end up doing topic modeling instead of frame analysis. 
Using a deductive approach, Dallas et al. \cite{dallas} created a dataset with articles about polemic topics such as immigration, same sex marriage, or smoking, and they defined 15 types of frames: "economic, capacity and resources, morality, fairness and equality, legality, constitutionality and jurisprudence, policy prescription and evaluation, crime and punishment, security and defense, health and safety, quality of life, cultural identity,
political, external regulation and reputation, other". In this case, they authors did not use a list of questions. Instead, for each article, annotators were asked to identify any of the 15 framing dimensions present in the article and to label text blurbs that cued them (based on the definitions of each of the frame dimensions) and decide the main frame of each article. In our case, we followed the idea of detecting the main frame by reading the text instead of answering questions, but instead of using the 15 frames proposed in \cite{dallas} , we used the 5 generic frames proposed in \cite{semetko}.


A final decision in our work was the type of text to analyze, whether headlines or whole article. For this decision, the chosen classification method was also going to be important. For example,
Khanehzar et al. \cite{Khanehzar2019ModelingPF} used traditional approaches such as SVMs as baseline, and demonstrated the improvement in frame classification with the use of pre-trained languages models such as BERT, RoBERTa and XLNet, following a fine-tuning approach, setting as input text a maximum of 256 tokens (although the maximum number of input tokens in these models is 512 tokens.)
Liu et al. \cite{liu2019detecting} classified news headlines about the gun problem in the United States, arguing for the choice of headlines as a unit of analysis based on previous journalism literature \cite{bleich2015media}, \cite{pan1993framing}, that advocated for the importance and influence of headlines on readers and the subsequent perception of articles.
From a computational viewpoint, using headlines is also an advantage, since you avoid the 512 token limitation in BERT-based models. Therefore, we decided to work with headlines about a controversial issue, namely the Covid-19 no-vax movement.

\begin{table}[!h]\centering
\caption{Summary of deductive approaches for frame analysis}\label{tab: }
\scriptsize
\resizebox{8.5cm}{!} {
\begin{tabular}{|p{0.15cm}|p{3.2cm}|p{2cm}|p{1.25cm}|p{1.25cm}|p{0.64cm}|}\toprule
\textbf{Ref} &\textbf{Frames} &\textbf{Goal} &\textbf{Technique} &\textbf{Number of samples} \\\midrule
\cite{dallas} &15 generic frames: "Economic", "Capacity and resources", "Morality", "Fairness and equality", "Legality, constitutionality and jurisprudence", "Policy prescription and evaluation", "Crime and punishment", "Security and defense", "Health and safety", "Quality of life", "Cultural identity", "Public opinion", "Political", "External regulation and reputation", "Other". &To label frames of full articles &Reading the full article, the annotator defines the main frame &~20000 articles \\ \hline
\cite{Khanehzar2019ModelingPF} &15 generic frames &Classification &BERT based models &~12000 articles \\ \hline
\cite{semetko} &5 generic frames: "human interest", "conflict", "morality", "attribution of responsibility", and "economic consequences". & To label frames of full articles &Yes/No questions.  &2600 articles and 1522 tv news stories \\ \hline

\cite{liu2019detecting} &9 specific frames:“Politics”, “Public opinion”, “Society/Culture”, and “Economic consequences” , “2nd Amendment” (Gun Rights), “Gun control/regulation”, “Mental health”, “School/Public space safety”, and “Race/Ethnicity”. &To label frames of full articles/ Classification &Reading the full article, the annotator defines the main frame. BERT based models &2990 headlines \\ \hline
\cite{el2021feverish} &5 generic frames + blame frame and fear frame &To label frames of full articles &Yes/No questions. &1170 articles \\ \hline
\cite{adiprasetio2020pandemic}&5 generic frames &To label frames of full articles &Reading the full article, the annotator defines the main frame. &6713 articles \\ \hline
\cite{rodelo2021framing} &5 generic frames + pandemic frames &To label frames of full articles &Yes/No questions. &2742 articles \\ \hline
\cite{catalan2019media} &5 generic frames, journalistic role and pandemic frames &To label frames of full articles &Reading the headline and subheadline, the annotator defines the main frame. &131 headlines + subheadlines \\\hline
\bottomrule
\end{tabular}
}
\end{table}

Continuing with the question of the methods used for classification, much work has been developed in prompt engineering, especially since the release of GPT-3.
Liu et al.\cite{liu2021pre} presented a good overview of the work done on this new NLP paradigm, not only explaining the concept of prompt engineering, but also the different strategies that can be followed both in the design of prompts, the potential applications, and the challenges to face when using
this approach. Prompt engineering applications include  knowledge probing \cite{qin2021learning}, information extraction \cite{shin2021constrained}, 
NLP reasoning \cite{trinh2018simple}, question answering \cite{jiang2020can},  text generation \cite{dou2020gsum}, multi-modal learning \cite{tsimpoukelli2021multimodal}, and text classification \cite{gao2020making}, the latter being the prompt-engineering use case in our work.
Puri et al.\cite{puri2019zero} presented a very interesting idea that we apply to our classification task. This consists of providing the language model with natural language descriptions of classification tasks as input, and training it to generate the correct answer in natural language via a language modeling objective. It is a zero-shot learning approach, in which no examples are used to explain the task to the model. Radford et al. \cite{radford2019language} demonstrated that language models can learn tasks without any explicit supervision. We have followed this approach to find an alternative way to do frame analysis.

As mentioned before, the emergence of  giant models like GPT-3, BLOOM, and ChatGPT are a very active research topic. To the best of our knowledge, on one hand our work extends the computational analysis of news related to the covid-19 no-vax movement, which illustrates the influence of the press on the ways societies think about relevant issues \cite{middleton2018social}, \cite{vannoy2010social}, and on the other hand it adds to the literature of human-machine interaction, regarding the design of GPT-3 prompts for classification tasks \cite{meyer2022we}, \cite{alex2021raft}.

\section{Data: European Covid-19 News Dataset}\label{section:data}



We used part of the European Covid-19 News dataset collected in our recent work \cite{how_european}. This dataset contains 51320 articles on Covid-19 vaccination from 19 newspapers from 5 different countries: Italy, France, Spain, Switzerland and UK. The articles cover a time period of 22 months, from January 2020 to October 2021. All content was translated into English to be able to work in a common language. The dataset was used for various analyses, such as name entity recognition, sentiment analysis, and subtopic modeling, to understand how Covid-19 vaccination was reported in Europe through the print media (in digital format.) The subtopic modeling analysis revealed a subsample of articles on the no-vax movement, which is the one we have used in this paper. We took the headlines of the articles associated with the no-vax movement, selecting all articles containing any of the keywords in Table \ref{tab:no_vax_keywords} in the headline or in the main text. This corresponds to a total of 1786 headlines.

\begin{table}[!htp]\centering
\caption{Keywords used to identify no-vax articles}\label{tab:no_vax_keywords}
\scriptsize
\resizebox{8.5cm}{!} {
\begin{tabular}{|cc|}\toprule
&\textbf{Keywords} \\\midrule
\textbf{NO VAX TOPIC} &"anti-vaxxers", "anti-vaccine", "anti-vaxx", "anti-corona", "no-vax", "no vax","anti-vaccin" \\
\bottomrule
\end{tabular}
}
\end{table}

In Table \ref{tab:headlines_country}, we show the number of headlines per country and newspaper. France is the country with the most no-vax articles in the corpus, with 523 articles, followed by Italy with 508. However, note that there are 6 newspapers from France, while only 2 from Italy.  Corriere della Sera is the newspaper that dealt most frequently with the subject (429 articles), while The Telegraph is the second one (206 articles). The total number of articles normalized by the number of newspapers per country is also shown in the last column of the Table. Using these normalized values, the ranking is Italy, UK, France, Switzerland, and Spain.

\begin{table}[!htp]\centering
\caption{Number of headlines by newspaper and country}\label{tab:headlines_country}
\scriptsize
\begin{tabular}{|c|c|c|c|c|}\toprule
\textbf{COUNTRY} &\textbf{NEWSPAPER} &\textbf{HEADLINES} &\textbf{TOTAL (NORM. TOTAL)} \\\midrule

{\textbf{FRANCE}} &La Croix &94 &{523 (87.1)} \\
&Le Monde &125 & \\
&Les Echos &49 & \\
&Liberation &97 & \\
&Lyon Capitale &8 & \\
&Ouest France &150 & \\ \hline
{\textbf{ITALY}} &Corriere della Sera &429 &{508 (254.0)} \\
&Il Sole 24 Ore &79 & \\ \hline
{\textbf{SPAIN}} &20 minutos &27 &{303 (50.5)} \\
&ABC &50 & \\
&El Diario &32 & \\
&El Mundo &77 & \\
&El Español &22 & \\
&La Vanguardia &95 & \\ \hline
{\textbf{SWITZERLAND}} &24 heures &97 &{230 (76.6)} \\
&La Liberté &22 & \\
&Le Temps &111 & \\ \hline
{\textbf{UNITED KINGDOM}} &The Irish News &16 &{222 (111.0)} \\
&The Telegraph &206 & \\ \hline
\textbf{} & &\textbf{} &\textbf{1786} \\ 
\bottomrule
\end{tabular}
\end{table}


\section{Methodology}\label{section:methodology}

\subsection{Human labeling of news frames}
To carry out the labeling of the frames in our corpus of headlines, we first designed a codebook, which contained the definitions of each of the frame types and a couple of examples of each type, as well as a definition of the corpus subject matter and definitions of the concept of frame analysis, so that the annotators could understand the task to be performed. The codebook follows the proposed by \cite{semetko} with 5 generic frames (attribution of responsibility, human interest, conflict, morality, and economic consequences) plus one additional 'no-frame' category. Two researchers were engaged to annotate a sample of the collected newspaper articles following a three-phase training procedure.

In the first phase, annotators had to read the codebook and get familiar with the task. In the second phase, they were asked to identify the main frame in the same subset of 50 headlines. At the end of the second phase, the intercoder reliability (ICR) was 0.58 between the 2 annotators. We analyzed those cases where there were discrepancies, and observed that in some cases, there was not a unique main frame, because both annotators had valid arguments to select one of the frames. In other cases, the discrepancies were due to slight misunderstanding of the definitions.
In the third phase, the annotators coded again 50 headlines, and the ICR increased to was 0.66. We realized that the possibility of having two frames remained. They discussed the cases in which they had disagreed, and if the other person's arguments were considered valid, it could be said that there were two frames.
After this three-phase training procedure, 
annotators were ready to annotate the dataset independently. 
We divided the dataset into two equal parts, and each person annotated 893 headlines.


\subsection{Fine-tuning GPT-3.5 and BERT-based models}

With the annotated dataset, we 
investigated two NLP approaches: the first one involves fine-tuning a pre-trained model; the second one is prompt engineering.
\begin{figure}[h]
  \centering
  \includegraphics[width=\linewidth]{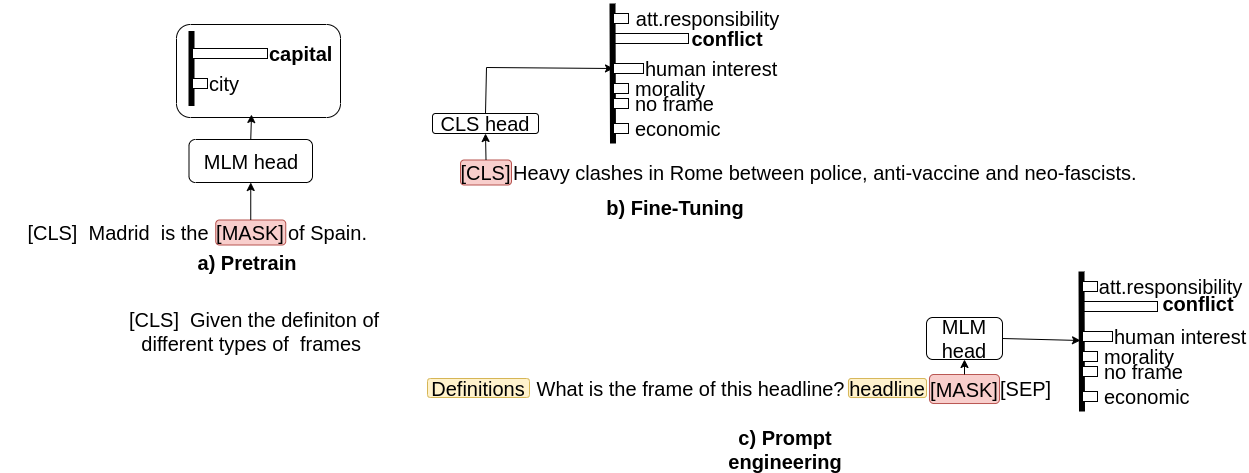}
  \caption{Pre-train, fine-tune, prompt}
  \label{fig:pretrain-finetuning-prompt}
  \Description{Number of headlines by newspaper and country.}
\end{figure}
Pre-trained language models have been trained with large text strings based on two unsupervised tasks, next sentence prediction and masked language model. Figure \ref{fig:pretrain-finetuning-prompt}  summarizes these techniques. 

In the first approach, a model with a fixed architecture is pre-trained as a language model (LM), predicting the likelihood of the observed textual data. This can be done due to the availability of large, raw text data needed to train LMs. This learning process can produce general purpose features of the modeled language. 
The learning process produces robust, general-purpose features of the language being modeled. 
The above pre-trained LM is then adapted to different downstream tasks, by introducing additional parameters and adjusting them using task-specific objective functions. In this approach, the focus was primarily on goal engineering, designing the training targets used in both the pre-training and the fine-tuning stages \cite{liu2021pre}.


We present an example to illustrate the idea. Imagine that the task is sentiment analysis, and we have a dataset with sentences and their associated sentiment, and a pre-trained model, which is a saved neural network trained with a much larger dataset. For that pre-trained model to address the target task, 
we unfreeze a few of the top layers of the saved model base and jointly train both the newly-added classifier layers and the last layers of the base model. This allows to "fine-tune" the higher-order feature representations in the base model to make them more relevant for the sentiment analysis task. In this way, instead of having to obtain a very large dataset with target labels to train a model, we can reuse the pre-trained model and use a much smaller train dataset. We use a part of our dataset as examples for the model to learn the task, while the other part of the dataset is used to evaluate model performance. 

Previous works related to frame classification in the computing literature have used fine-tuning, BERT-based models. In our work, we have done the same as a baseline, but we aimed to go one step further and also produce results using fine-tuning of GPT-3.5.

\subsection{Prompt-engineering with GPT-3.5}
Model fine-tuning has been widely used, but with the emergence of generative models such as GPT-3, another way to approach 
classification tasks has appeared. The idea is to use the pre-trained model directly and convert the task to be performed into a format as close as possible to the tasks for which it has been pre-trained.
That is, if the model has been pre-trained from next word prediction as in the case of GPT-3, 
classification can be done by defining a prompt, where the input to the model is an incomplete sentence, and the model must complete it with a word or several words, just as it has been trained. 
This avoids having to use part of the already labeled dataset to teach the task to be performed to the model, and a previous labeling is not needed 
\cite{liu2021pre}.

In this approach, instead of adapting pre-trained LMs to downstream tasks via objective engineering, downstream tasks are reformulated to look more like those solved during the original LM training with the help of a textual prompt. For example, when recognizing the emotion of a social media post, “I missed the bus today.”, we may continue with a prompt “I felt so
\textunderscore”, and ask the LM to fill the blank with an emotion-bearing word. Or if we choose the prompt “English: I missed the bus today. French:
\textunderscore”), an LM may be able to fill in the blank with a French translation. In this way, by selecting the appropriate prompts, we can influence the model behavior so that the pre-trained LM itself can be used to predict the desired output, even without any additional task-specific training \cite{liu2021pre}.

\begin{figure}[h]
  \centering
  \includegraphics[width=\linewidth]{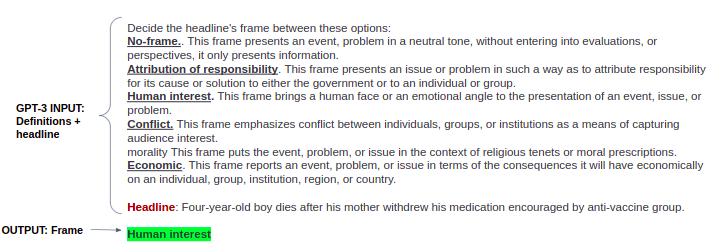}
  \caption{GPT-3.5 for frame inference: input and output}
  \label{fig:gpt3_input}
\end{figure}

We use this emerging NLP approach to classify frames at headline level. 
We are not aware of previous uses of this strategy to classify frames as we propose here. The idea is the following. Prompt engineering consists of giving a prompt to the model, and understands that prompt as an incomplete sentence.
To do prompt engineering with our dataset, we needed to define an appropriate prompt 
that would produce the headline frames as output.
We defined several experiments with the Playground of GPT-3, in order to find the best prompt for our task. In our initial experiments, we followed existing approaches in prompt engineering to do sentiment analysis, where the individual answer was an adjective, and this adjective was matched with a sentiment. In a similar fashion, we decided to build a thesaurus of adjectives that define each of the frames. For instance, the human interest frame could be 'interesting', 'emotional', 'personal', 'human'. The conflict frame could be: 'conflictive', 'bellicose', 'troublesome', 'rowdy', 'quarrelsome', 'troublemaker', 'agitator', etc. After the list of adjectives was defined, we needed to define the prompt in order to get, as an answer, one of the adjectives in our thesaurus to match them with the frame. We used the GPT-3 playground using the headline as input and asking for the frame as output, but the strategy did not work. In our final experiment, instead of giving the headline as input, we gave the definitions of each type of frame plus the headline, and we asked the model to choose between the different types of frames as output. In this way, the output of the model was directly one of the frames, and we avoided the step of matching adjectives with  frames. An example is shown in Figure \ref{fig:gpt3_input}. 

For the GPT-3 configuration \footnote{https://beta.openai.com/docs/introduction}, there are 3 main concepts:
\begin{itemize}
    \item TEMPERATURE [0-1]. This parameter controls randomness, lowering it results in less random completions.
    \item TOP\_P [0-1]. This parameter controls diversity via nucleus sampling.
    \item MAX\_TOKENS[1-4000]. This parameter indicates the maximum number of tokens to generate,
    \item MODEL. GPT-3 offer four main models with different levels of power, suitable for different tasks. Davinci is the most capable model, and Ada is the fastest.
    
\end{itemize}
After testing with the GPT-3 playground and varying different hyper-parameters to assess performance, we set the temperature to 0, since the higher the temperature the more random the response. Furthermore, the Top-p parameter was set to 1, as it would likely get a set of the most likely words for the model to choose from. The maximum number of tokens was set to 2; in this way, the model is asked to choose between one of the responses. As a model, we used the one with the best performance at the time of experimental design, which was TEXT-DAVINCI-003, recognized as GPT 3.5.

\section{Results: Human Labeling of Frames in No-Vax News Headlines (RQ1)}\label{section:results_discussion}


In this section, we present and discuss the results of the analysis related to our first RQ.

\begin{figure}[h]
  \centering
  \includegraphics[width=\linewidth]{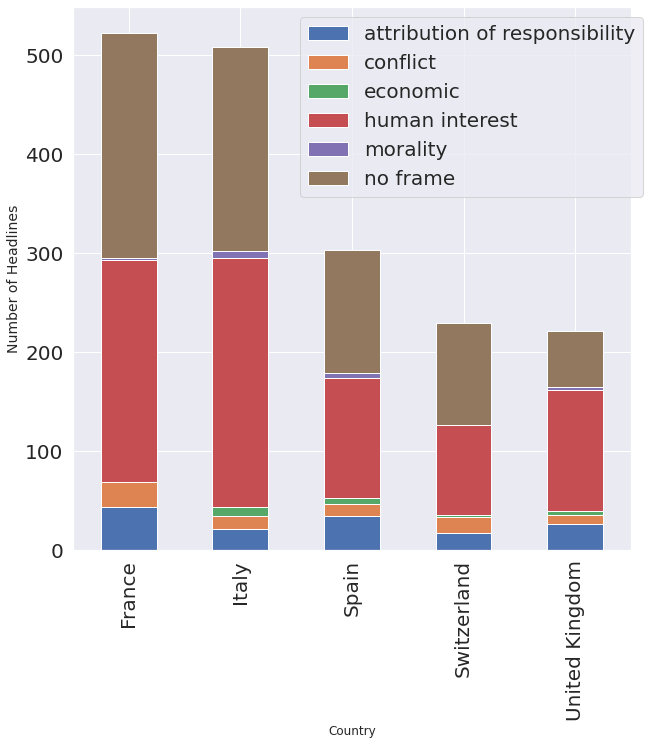}
  \caption{Non-normalized distribution of frames per country}
  \label{fig:frames_country}
  \Description{Number of headlines by newspaper and country.}
\end{figure}

Figure \ref{fig:frames_country} shows the distribution of frames per country at headline level, with human interest and no-frame being the predominant ones. Attribution of responsibility is the third one except in Switzerland, where the corresponding frame is conflict. Finally,  morality and economic are the least represented in the dataset for every country.

\begin{figure}[h]
  \centering
  \includegraphics[width=\linewidth]{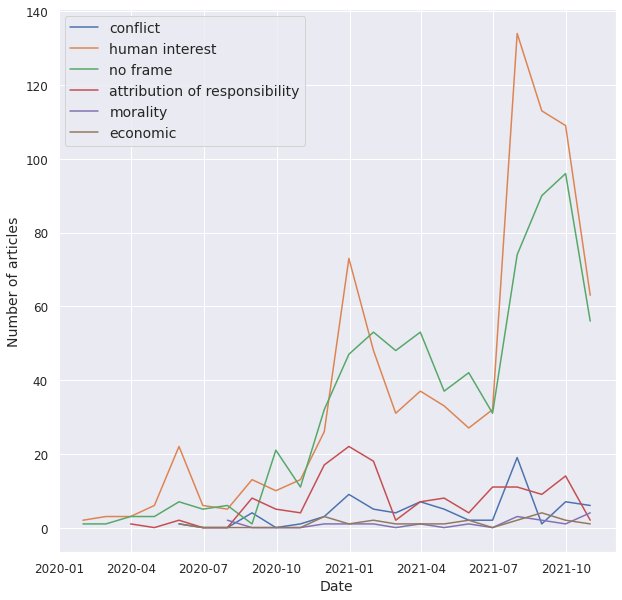}
  \caption{Non-normalized monthly distribution of frames.}
  \label{fig:frames_date}
  \Description{frames over time.}
\end{figure}

The monthly distribution of frames aggregated for all countries is shown in Fig. \ref{fig:frames_date}.
We can see two big peaks, the first one in January 2021 and the second one in August 2021. In all countries, the vaccination process started at the end of December 2020, so it makes sense that the no-vax movement started to be more predominant in the news in January 2021. Human interest is the most predominant frame. Manual inspection shows that this is because the headlines are about personal cases of people who are pro- or anti- vaccine. Attribution of responsibility is also present. Manual inspection indicates that local politicians and health authorities had to make decisions about who could be vaccinated at the beginning of the process. The second peak at the end of summer 2021 coincided with the health pass (also called Covid passport in some countries), and we can observe a peak in the curve corresponding to the conflict frame, reflecting the demonstrations against the measure of mandatory health passes taken by country governments.

\begin{figure}[h]
  \centering
  \includegraphics[width=\linewidth]{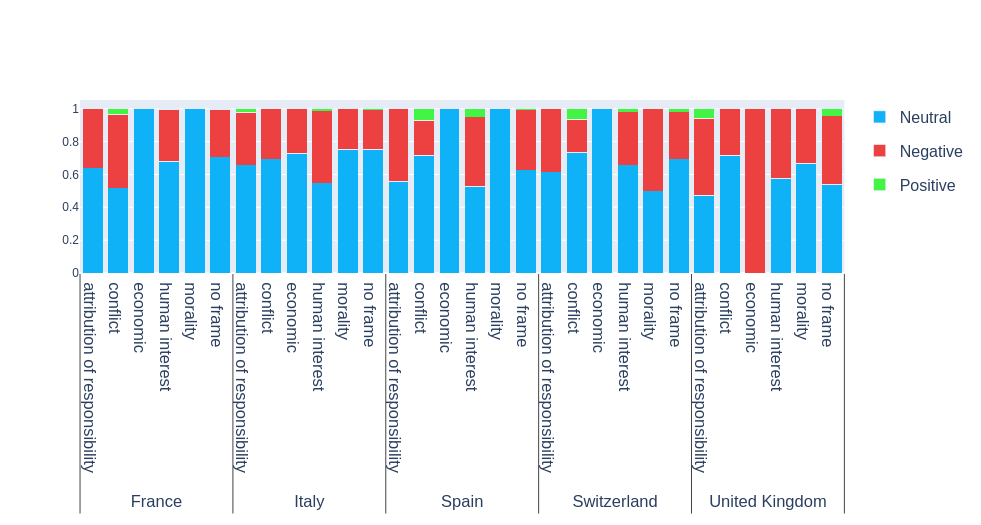}
  \caption{Sentiment of headline by frame and by country}
  \label{fig:sentiment}
  \Description{Number of headlines by newspaper and country.}
\end{figure}

In Figure \ref{fig:sentiment}, we compare the sentiment per frame and per country, to understand if there were any major differences. The sentiment analysis labels were obtained using BERT-sent from the Hugging Face package \cite{bertsent}, used in our previous work (please refer to our original analysis in \cite{how_european} for details.)
We normalized the results between 0 and 1 to compare frames between countries. We see that the sentiment is predominantly neutral (in blue). Examining in more detail the negative and positive sentiment of each frame category, we observed a few trends:
\begin{itemize}
\item Attribution of responsibility: Negative sentiment represents 30-40\% of the cases, while positive tone is only found in residual form in Italy, Switzerland, and the United Kingdom.
\item Conflict: Negative sentiment represents 20-35\% of the cases.
\item  Economic: Predominantly neutral, with only negative tone in Italy and UK (in the latter case, all headlines with this frame were considered negative.)
\item Human interest: Negative sentiment represents 30-40\% of the cases, while positive tone is only found in residual form in Italy, Spain, and Switzerland.
\item Morality: Predominantly neutral, with negative tone in Italy, Switzerland, and the United Kingdom,
\item No frame: 20-30\% of negative content.
\end{itemize}

Regarding the results of the annotation process, the fact that the distribution of the 6 frame types is relatively similar between countries 
suggests that the anti-vaccine movement issue was treated in a similar way in these countries. The fact that human interest is the most dominant frame indicates 
that this issue was treated from a more human and emotional approach, with headlines about personal experiences, celebrities giving their opinion about vaccination, and politicians defending vaccine policies. Moreover, the reason for many headlines being classified as no-frame is partly due to how data was selected. We chose articles that contained words related to no-vax, either in the headline or in the article. This resulted in many headlines not containing anything specific related to no-vax, while the no-vax content was actually included in the main text of the corresponding articles.


It is worth mentioning that prior to obtaining the results, we had expected that attribution of responsibility would be among the most prominent frames, since governments took many measures such as mandatory health pass requirements to access certain sites; we had also expected that the conflict frame would be prominent, since there were many demonstrations in Europe. In reality, however, these frames categories were not reflected as frequently at the headline level.

Regarding the analysis at the temporal level, it is clear that 
certain events were captured by the press, such as the start of vaccination or the mandatory vaccination passport.

Finally, the sentiment analysis of the different frames shows that the predominant tone in all of them is neutral or negative, with very similar trends between countries. This association between sentiment analysis and frames has been discussed in previous literature \cite{burscher2016frames} \cite{nicholls2021computational}.

\section{Results: GPT-3.5 for frame classification of headlines (RQ2)}\label{section:results_discussion2}


Here, we present and discuss the results  related to our second RQ.

\subsection{Fine-tuning GPT-3.5}

Table \ref{tab:cross_validation} shows the results of the 6-class classification task using 5-cross validation. Three models were used: GPT-3.5 and two BERT-based models. We observe that, on average, GPT-3.5 performs  better than the BERT-based models. This is somehow expected as GPT-3.5 is a much larger model. 
Overall, in the case of fine-tuning, the best performance for the six-class frame classification task is 72\% accuracy, which is promising, with an improvement over previous models based on BERT. Yet, it should be noted that the performance differences are modest (2\% improvement between GPT-3.5 and RoBERTa). 

\begin{table}[!htp]\centering
\caption{Classification results for six-class frame classification and 5-fold cross validation}\label{tab:cross_validation}
\scriptsize
\resizebox{8.5cm}{!} {
\begin{tabular}{|lcccccc|}\toprule
\textbf{ACCURACY} &\textbf{0} &\textbf{1} &\textbf{2} &\textbf{3} &\textbf{4} &\textbf{AVERAGE} \\\midrule
\textbf{BERT} &0.68 &0.69 &0.72 &0.64 &0.70 &0.67 \\
\textbf{RoBERTa} &0.70 &0.72 &0.72 &0.67 &0.71 &0.70 \\
\textbf{GPT3} &0.75 &0.70 &0.72 &0.71 &0.71 &0.72 \\
\bottomrule
\end{tabular}
}
\end{table}

On the other hand, BERT is open-source, while GPT-3 has an economic cost as the use of the model is not free, which monetarily limits the number of experiments that can be performed with it, as well as the different configurations one can explore to improve performance. This is important because much of the improvement in performance requires empirical explorations of model parameters 
More specifically, the cost of an experiment for each of the folds has a cost of 4 dollars (at the time of writing this paper.)
This represents a limitation in practice. 

Furthermore, GPT-3 has a significant carbon footprint. Similarly, for prompt engineering (discussed in the next subsection), choosing the right prompt (i.e., the words that best define the task so that the model is able to perform adequately) is also based on trial and error. This also has an impact on carbon footprint.
In connection with this topic, Strubell et al.\cite{strubell2019energy} argue that improvements in the accuracy of models depend on the availability of large computational resources, which involve large economic and environmental costs.  A criticism has been made as 'the rich get richer', in the sense that not all research groups have sufficient infrastructure resources and access to funding needed to use these models and improve their performance.
Also in relation to this analysis, the work of Bender et al. \cite{bender2021dangers}  evaluates the costs and risks of the use of large language models, stating that researchers should be aware of the impact that these models have on the environment, and assess whether the benefits outweigh the risks. The work in \cite{bender2021dangers} provides a very telling example, where people living in the Maldives or Sudan are affected by floods and pay the environmental price of training English LLMs, when similar models have not been produced for languages like Dhivehi or Sudanese Arab. In short, there is a need to establish ways to use this technological development responsibly, and it all starts with being aware of the risks it presents.


\subsection{Prompt-engineering with GPT-3.5}
For each headline, we got the frame that the model considered the most likely, and we compared these GPT-3.5 inferences with the frames labeled by the annotators. The agreement between model and annotator was of 49\%.
Analyzing the results, and specifically looking at the cases where the annotator and GPT-3.5 disagreed, we discovered that according to the frame definitions, the model in some cases proposed a frame that indeed made sense. This observation, together with our previous experience in the annotation process, where headlines could have more than one valid frame, led us to design a second post-hoc experiment.
We took all the headlines where each of the two annotators had disagreed with GPT-3.5, and 
we asked the annotators to state whether they would agree (or not) with each GPT-inferred label for a given headline. It is important to emphasize that the annotators did not know the origin of that label, i.e., they did not know if it was the label they had originally assigned, or if it was a random one. In this way, we could quantify how GPT-3.5 worked according to valid arguments provided by the annotators. In this post-hoc experiment, the model agreed in 76\% of cases with the annotators.


Looking at the results of the classification models, the 49\% accuracy of the prompt-engineering approach can be considered low, yet we consider that it is a valid avenue for further investigation, as in the second post-hoc analysis, we found that the model agrees with human annotators in 76\% of the cases. Clearly, framing involves aspects of subjectivity \cite{munro2021human}. Much of what we do as people has a subjective component, influenced by how we feel or how we express opinions.  


News reading is never fully objective, and the annotators engaged in the frame classification task, influenced by their personal state of mind, experience, and culture, may perceive information differently. Monarch affirms that "for simple tasks, like binary labels on objective tasks, the statistics are fairly straightforward to decide which is the ‘correct’ label when different annotators disagree. But for subjective tasks, or even objective tasks with continuous data, there are no simple heuristics for deciding what the correct label should be" \cite{munro2021human}.

Subjectivity is involved in both the generation and perception of information: the assumption that there is only one frame is complicated by the point of view of the reader. In the case of news, the information sender (the journalist) has an intention, but the receiver (the reader) plays a role and is influenced by it. In psychology, this is known as the lens model of interpersonal communication, where the sender has certain objectives, but the receiver can interpret or re-interpret what the sender wants to say, with more or less accuracy \cite{lens}. 

Following this discussion on subjectivity, the question arose as to what would happen if, instead of headlines, we used the complete article as a source of analysis. We wondered if longer text could make the frame labeling task clearer than when using headlines. 
Yet another possible hypothesis is that having to read longer texts could lead to the same subject being presented from different angles. Please recall that in the existing literature discussed in Section 2, both headlines and full articles have been used from frame analysis (see Table 1.) This remains as an issue for future work.

\section{Conclusions}\label{section:conclusions}
In this paper, we first presented an analysis of human-generated news frames on the covid-19 no-vax movement in Europe, and then studied different approaches using large language models for automatic inference of frames. We conclude by answering the two research questions we posed:

RQ1: What are the main frames in the news headlines about the covid-19 anti-vaccine movement in 5 European countries?
After annotating the headlines, we found that of the 1786 headlines, the predominant frame is human interest (45.3\% of cases),
which presents a news item with an emotional angle, putting a face to a problem or situation. We also found that a substantial proportion of headlines were annotated as not presenting any frame (40.2\% of cases). Finally, the other frame types are found more infrequently.

RQ2: Can prompt engineering be used for classification of headlines according to frames?
We first used fine-tuning of a number of language models, and found that GPT-3.5 produced classification accuracy of 72\% on a six-frame classification task. This represented a modest 2\% improvement over BERT-based models, at a significantly larger environmental cost. We then presented a new way of classifying frames using prompts. At the headline level, inferences made with GPT-3.5 reached  49\% of agreement with human-generated frame labels. In many cases, the GPT-3.5 model inferred frame types that were considered as valid choices by human annotators, and in an post-doc experiment, the human-machine agreement reached 76\%. These results have opened several new directions for future work.


 


\begin{acks}
This work was supported by the AI4Media project, funded by the European Commission (Grant 951911) under the H2020 Programme ICT-48-2020. We also thank the newspapers for sharing their online articles. Finally, we thank our colleagues Haeeun Kim and Emma Bouton-Bessac for their support with annotations, and Victor Bros and Oleksii Polegkyi for discussions.
\end{acks}




\bibliographystyle{ACM-Reference-Format}
\bibliography{sample-base}


\begin{thebibliography}{60}


\ifx \showCODEN    \undefined \def \showCODEN     #1{\unskip}     \fi
\ifx \showDOI      \undefined \def \showDOI       #1{#1}\fi
\ifx \showISBNx    \undefined \def \showISBNx     #1{\unskip}     \fi
\ifx \showISBNxiii \undefined \def \showISBNxiii  #1{\unskip}     \fi
\ifx \showISSN     \undefined \def \showISSN      #1{\unskip}     \fi
\ifx \showLCCN     \undefined \def \showLCCN      #1{\unskip}     \fi
\ifx \shownote     \undefined \def \shownote      #1{#1}          \fi
\ifx \showarticletitle \undefined \def \showarticletitle #1{#1}   \fi
\ifx \showURL      \undefined \def \showURL       {\relax}        \fi
\providecommand\bibfield[2]{#2}
\providecommand\bibinfo[2]{#2}
\providecommand\natexlab[1]{#1}
\providecommand\showeprint[2][]{arXiv:#2}

\bibitem[Adiprasetio and Larasati(2020)]%
        {adiprasetio2020pandemic}
\bibfield{author}{\bibinfo{person}{Justito Adiprasetio} {and}
  \bibinfo{person}{Annissa~Winda Larasati}.} \bibinfo{year}{2020}\natexlab{}.
\newblock \showarticletitle{Pandemic crisis in online media: Quantitative
  framing analysis on Detik. com’s coverage of Covid-19}.
\newblock \bibinfo{journal}{\emph{Jurnal Ilmu Sosial Dan Ilmu Politik}}
  \bibinfo{volume}{24}, \bibinfo{number}{2} (\bibinfo{year}{2020}),
  \bibinfo{pages}{153--170}.
\newblock


\bibitem[Alex et~al\mbox{.}(2021)]%
        {alex2021raft}
\bibfield{author}{\bibinfo{person}{Neel Alex}, \bibinfo{person}{Eli Lifland},
  \bibinfo{person}{Lewis Tunstall}, \bibinfo{person}{Abhishek Thakur},
  \bibinfo{person}{Pegah Maham}, \bibinfo{person}{C~Jess Riedel},
  \bibinfo{person}{Emmie Hine}, \bibinfo{person}{Carolyn Ashurst},
  \bibinfo{person}{Paul Sedille}, \bibinfo{person}{Alexis Carlier},
  {et~al\mbox{.}}} \bibinfo{year}{2021}\natexlab{}.
\newblock \showarticletitle{RAFT: A real-world few-shot text classification
  benchmark}.
\newblock \bibinfo{journal}{\emph{arXiv preprint arXiv:2109.14076}}
  (\bibinfo{year}{2021}).
\newblock


\bibitem[Alonso~del Barrio and Gatica-Perez(2022)]%
        {how_european}
\bibfield{author}{\bibinfo{person}{David Alonso~del Barrio} {and}
  \bibinfo{person}{Daniel Gatica-Perez}.} \bibinfo{year}{2022}\natexlab{}.
\newblock \showarticletitle{How Did Europe's Press Cover Covid-19 Vaccination
  News? A Five-Country Analysis}.
\newblock  (\bibinfo{year}{2022}), \bibinfo{pages}{35–43}.
\newblock
\showISBNx{9781450392426}
\urldef\tempurl%
\url{https://doi.org/10.1145/3512732.3533588}
\showDOI{\tempurl}


\bibitem[Bender et~al\mbox{.}(2021)]%
        {bender2021dangers}
\bibfield{author}{\bibinfo{person}{Emily~M Bender}, \bibinfo{person}{Timnit
  Gebru}, \bibinfo{person}{Angelina McMillan-Major}, {and}
  \bibinfo{person}{Shmargaret Shmitchell}.} \bibinfo{year}{2021}\natexlab{}.
\newblock \showarticletitle{On the Dangers of Stochastic Parrots: Can Language
  Models Be Too Big?}
\newblock  (\bibinfo{year}{2021}), \bibinfo{pages}{610--623}.
\newblock


\bibitem[Biswal and Gouda(2020)]%
        {biswal2020artificial}
\bibfield{author}{\bibinfo{person}{Santosh~Kumar Biswal} {and}
  \bibinfo{person}{Nikhil~Kumar Gouda}.} \bibinfo{year}{2020}\natexlab{}.
\newblock \showarticletitle{Artificial intelligence in journalism: A boon or
  bane?}
\newblock In \bibinfo{booktitle}{\emph{Optimization in machine learning and
  applications}}. \bibinfo{publisher}{Springer}, \bibinfo{pages}{155--167}.
\newblock


\bibitem[Bleich et~al\mbox{.}(2015)]%
        {bleich2015media}
\bibfield{author}{\bibinfo{person}{Erik Bleich}, \bibinfo{person}{Hannah
  Stonebraker}, \bibinfo{person}{Hasher Nisar}, {and} \bibinfo{person}{Rana
  Abdelhamid}.} \bibinfo{year}{2015}\natexlab{}.
\newblock \showarticletitle{Media portrayals of minorities: Muslims in British
  newspaper headlines, 2001--2012}.
\newblock \bibinfo{journal}{\emph{Journal of Ethnic and Migration Studies}}
  \bibinfo{volume}{41}, \bibinfo{number}{6} (\bibinfo{year}{2015}),
  \bibinfo{pages}{942--962}.
\newblock


\bibitem[Bommarito and Katz(2022)]%
        {gpt_bar_exam}
\bibfield{author}{\bibinfo{person}{Michael Bommarito} {and}
  \bibinfo{person}{Daniel~Martin Katz}.} \bibinfo{year}{2022}\natexlab{}.
\newblock \bibinfo{title}{GPT Takes the Bar Exam}.
\newblock
\newblock
\urldef\tempurl%
\url{https://doi.org/10.48550/ARXIV.2212.14402}
\showDOI{\tempurl}


\bibitem[Broussard et~al\mbox{.}(2019)]%
        {broussard2019artificial}
\bibfield{author}{\bibinfo{person}{Meredith Broussard},
  \bibinfo{person}{Nicholas Diakopoulos}, \bibinfo{person}{Andrea~L Guzman},
  \bibinfo{person}{Rediet Abebe}, \bibinfo{person}{Michel Dupagne}, {and}
  \bibinfo{person}{Ching-Hua Chuan}.} \bibinfo{year}{2019}\natexlab{}.
\newblock \showarticletitle{Artificial intelligence and journalism}.
\newblock \bibinfo{journal}{\emph{Journalism \& Mass Communication Quarterly}}
  \bibinfo{volume}{96}, \bibinfo{number}{3} (\bibinfo{year}{2019}),
  \bibinfo{pages}{673--695}.
\newblock


\bibitem[Brown et~al\mbox{.}(2020)]%
        {brown2020language}
\bibfield{author}{\bibinfo{person}{Tom Brown}, \bibinfo{person}{Benjamin Mann},
  \bibinfo{person}{Nick Ryder}, \bibinfo{person}{Melanie Subbiah},
  \bibinfo{person}{Jared~D Kaplan}, \bibinfo{person}{Prafulla Dhariwal},
  \bibinfo{person}{Arvind Neelakantan}, \bibinfo{person}{Pranav Shyam},
  \bibinfo{person}{Girish Sastry}, \bibinfo{person}{Amanda Askell},
  {et~al\mbox{.}}} \bibinfo{year}{2020}\natexlab{}.
\newblock \showarticletitle{Language models are few-shot learners}.
\newblock \bibinfo{journal}{\emph{Advances in neural information processing
  systems}}  \bibinfo{volume}{33} (\bibinfo{year}{2020}),
  \bibinfo{pages}{1877--1901}.
\newblock


\bibitem[Burscher et~al\mbox{.}(2014)]%
        {burscher2014teaching}
\bibfield{author}{\bibinfo{person}{Bj{\"o}rn Burscher}, \bibinfo{person}{Daan
  Odijk}, \bibinfo{person}{Rens Vliegenthart}, \bibinfo{person}{Maarten
  De~Rijke}, {and} \bibinfo{person}{Claes~H De~Vreese}.}
  \bibinfo{year}{2014}\natexlab{}.
\newblock \showarticletitle{Teaching the computer to code frames in news:
  Comparing two supervised machine learning approaches to frame analysis}.
\newblock \bibinfo{journal}{\emph{Communication Methods and Measures}}
  \bibinfo{volume}{8}, \bibinfo{number}{3} (\bibinfo{year}{2014}),
  \bibinfo{pages}{190--206}.
\newblock


\bibitem[Burscher et~al\mbox{.}(2016)]%
        {burscher2016frames}
\bibfield{author}{\bibinfo{person}{Bjorn Burscher}, \bibinfo{person}{Rens
  Vliegenthart}, {and} \bibinfo{person}{Claes H~de Vreese}.}
  \bibinfo{year}{2016}\natexlab{}.
\newblock \showarticletitle{Frames beyond words: Applying cluster and sentiment
  analysis to news coverage of the nuclear power issue}.
\newblock \bibinfo{journal}{\emph{Social Science Computer Review}}
  \bibinfo{volume}{34}, \bibinfo{number}{5} (\bibinfo{year}{2016}),
  \bibinfo{pages}{530--545}.
\newblock


\bibitem[Card et~al\mbox{.}(2015)]%
        {dallas}
\bibfield{author}{\bibinfo{person}{Dallas Card}, \bibinfo{person}{Amber
  Boydstun}, \bibinfo{person}{Justin Gross}, \bibinfo{person}{Philip Resnik},
  {and} \bibinfo{person}{Noah Smith}.} \bibinfo{year}{2015}\natexlab{}.
\newblock \showarticletitle{The Media Frames Corpus: Annotations of Frames
  Across Issues}.
\newblock   \bibinfo{volume}{2} (\bibinfo{date}{01} \bibinfo{year}{2015}),
  \bibinfo{pages}{438--444}.
\newblock
\urldef\tempurl%
\url{https://doi.org/10.3115/v1/P15-2072}
\showDOI{\tempurl}


\bibitem[Catalan-Matamoros and El{\'\i}as(2020)]%
        {catalan2020vaccine}
\bibfield{author}{\bibinfo{person}{Daniel Catalan-Matamoros} {and}
  \bibinfo{person}{Carlos El{\'\i}as}.} \bibinfo{year}{2020}\natexlab{}.
\newblock \showarticletitle{Vaccine hesitancy in the age of coronavirus and
  fake news: analysis of journalistic sources in the Spanish quality press}.
\newblock \bibinfo{journal}{\emph{International Journal of Environmental
  Research and Public Health}} \bibinfo{volume}{17}, \bibinfo{number}{21}
  (\bibinfo{year}{2020}), \bibinfo{pages}{8136}.
\newblock


\bibitem[Catal{\'a}n-Matamoros and Pe{\~n}afiel-Saiz(2019)]%
        {catalan2019media}
\bibfield{author}{\bibinfo{person}{Daniel Catal{\'a}n-Matamoros} {and}
  \bibinfo{person}{Carmen Pe{\~n}afiel-Saiz}.} \bibinfo{year}{2019}\natexlab{}.
\newblock \showarticletitle{Media and mistrust of vaccines: a content analysis
  of press headlines}.
\newblock \bibinfo{journal}{\emph{Revista latina de comunicaci{\'o}n social}}
  \bibinfo{number}{74} (\bibinfo{year}{2019}), \bibinfo{pages}{786--802}.
\newblock


\bibitem[Coddington(2015)]%
        {coddington2015clarifying}
\bibfield{author}{\bibinfo{person}{Mark Coddington}.}
  \bibinfo{year}{2015}\natexlab{}.
\newblock \showarticletitle{Clarifying journalism’s quantitative turn: A
  typology for evaluating data journalism, computational journalism, and
  computer-assisted reporting}.
\newblock \bibinfo{journal}{\emph{Digital journalism}} \bibinfo{volume}{3},
  \bibinfo{number}{3} (\bibinfo{year}{2015}), \bibinfo{pages}{331--348}.
\newblock


\bibitem[Cooper(2010)]%
        {cooper2010oppositional}
\bibfield{author}{\bibinfo{person}{Stephen~D Cooper}.}
  \bibinfo{year}{2010}\natexlab{}.
\newblock \showarticletitle{The oppositional framing of bloggers}.
\newblock In \bibinfo{booktitle}{\emph{Doing News Framing Analysis}}.
  \bibinfo{publisher}{Routledge}, \bibinfo{pages}{151--172}.
\newblock


\bibitem[Dale(2021)]%
        {dale2021gpt}
\bibfield{author}{\bibinfo{person}{Robert Dale}.}
  \bibinfo{year}{2021}\natexlab{}.
\newblock \showarticletitle{GPT-3: What’s it good for?}
\newblock \bibinfo{journal}{\emph{Natural Language Engineering}}
  \bibinfo{volume}{27}, \bibinfo{number}{1} (\bibinfo{year}{2021}),
  \bibinfo{pages}{113--118}.
\newblock


\bibitem[Dirikx and Gelders(2010a)]%
        {climate_change_frame}
\bibfield{author}{\bibinfo{person}{Astrid Dirikx} {and} \bibinfo{person}{Dave
  Gelders}.} \bibinfo{year}{2010}\natexlab{a}.
\newblock \showarticletitle{To frame is to explain: A deductive frame-analysis
  of Dutch and French climate change coverage during the annual UN Conferences
  of the Parties}.
\newblock \bibinfo{journal}{\emph{Public Understanding of Science}}
  \bibinfo{volume}{19}, \bibinfo{number}{6} (\bibinfo{year}{2010}),
  \bibinfo{pages}{732--742}.
\newblock
\urldef\tempurl%
\url{https://doi.org/10.1177/0963662509352044}
\showDOI{\tempurl}
\showeprint{https://doi.org/10.1177/0963662509352044}
\newblock
\shownote{PMID: 21560546}.


\bibitem[Dirikx and Gelders(2010b)]%
        {dirikx2010frame}
\bibfield{author}{\bibinfo{person}{Astrid Dirikx} {and} \bibinfo{person}{Dave
  Gelders}.} \bibinfo{year}{2010}\natexlab{b}.
\newblock \showarticletitle{To frame is to explain: A deductive frame-analysis
  of Dutch and French climate change coverage during the annual UN Conferences
  of the Parties}.
\newblock \bibinfo{journal}{\emph{Public understanding of science}}
  \bibinfo{volume}{19}, \bibinfo{number}{6} (\bibinfo{year}{2010}),
  \bibinfo{pages}{732--742}.
\newblock


\bibitem[Dou et~al\mbox{.}(2020)]%
        {dou2020gsum}
\bibfield{author}{\bibinfo{person}{Zi-Yi Dou}, \bibinfo{person}{Pengfei Liu},
  \bibinfo{person}{Hiroaki Hayashi}, \bibinfo{person}{Zhengbao Jiang}, {and}
  \bibinfo{person}{Graham Neubig}.} \bibinfo{year}{2020}\natexlab{}.
\newblock \showarticletitle{Gsum: A general framework for guided neural
  abstractive summarization}.
\newblock \bibinfo{journal}{\emph{arXiv preprint arXiv:2010.08014}}
  (\bibinfo{year}{2020}).
\newblock


\bibitem[Ebrahim(2022)]%
        {ebrahim2022corona}
\bibfield{author}{\bibinfo{person}{Sumayya Ebrahim}.}
  \bibinfo{year}{2022}\natexlab{}.
\newblock \showarticletitle{The corona chronicles: Framing analysis of online
  news headlines of the COVID-19 pandemic in Italy, USA and South Africa}.
\newblock \bibinfo{journal}{\emph{Health SA Gesondheid (Online)}}
  \bibinfo{volume}{27} (\bibinfo{year}{2022}), \bibinfo{pages}{1--8}.
\newblock


\bibitem[El-Behary(2021)]%
        {el2021feverish}
\bibfield{author}{\bibinfo{person}{Hend Abdelgaber~Ahmed El-Behary}.}
  \bibinfo{year}{2021}\natexlab{}.
\newblock \showarticletitle{A Feverish Spring: A Comparative Analysis of
  COVID-19 News Framing in Sweden, the UK, and Egypt}.
\newblock  (\bibinfo{year}{2021}).
\newblock


\bibitem[Entman(1993)]%
        {entman1993framing}
\bibfield{author}{\bibinfo{person}{Robert~M Entman}.}
  \bibinfo{year}{1993}\natexlab{}.
\newblock \showarticletitle{Framing: Towards clarification of a fractured
  paradigm}.
\newblock \bibinfo{journal}{\emph{McQuail's reader in mass communication
  theory}}  \bibinfo{volume}{390} (\bibinfo{year}{1993}), \bibinfo{pages}{397}.
\newblock


\bibitem[Gao et~al\mbox{.}(2020)]%
        {gao2020making}
\bibfield{author}{\bibinfo{person}{Tianyu Gao}, \bibinfo{person}{Adam Fisch},
  {and} \bibinfo{person}{Danqi Chen}.} \bibinfo{year}{2020}\natexlab{}.
\newblock \showarticletitle{Making pre-trained language models better few-shot
  learners}.
\newblock \bibinfo{journal}{\emph{arXiv preprint arXiv:2012.15723}}
  (\bibinfo{year}{2020}).
\newblock


\bibitem[Ghasiya and Okamura(2021)]%
        {ghasiya2021investigating}
\bibfield{author}{\bibinfo{person}{Piyush Ghasiya} {and} \bibinfo{person}{Koji
  Okamura}.} \bibinfo{year}{2021}\natexlab{}.
\newblock \showarticletitle{Investigating COVID-19 news across four nations: a
  topic modeling and sentiment analysis approach}.
\newblock \bibinfo{journal}{\emph{Ieee Access}}  \bibinfo{volume}{9}
  (\bibinfo{year}{2021}), \bibinfo{pages}{36645--36656}.
\newblock


\bibitem[Gifford(1994)]%
        {lens}
\bibfield{author}{\bibinfo{person}{Robert Gifford}.}
  \bibinfo{year}{1994}\natexlab{}.
\newblock \showarticletitle{A Lens-Mapping Framework for Understanding the
  Encoding and Decoding of Interpersonal Dispositions in Nonverbal Behavior}.
\newblock \bibinfo{journal}{\emph{Journal of Personality and Social
  Psychology}}  \bibinfo{volume}{66} (\bibinfo{date}{02} \bibinfo{year}{1994}),
  \bibinfo{pages}{398--412}.
\newblock
\urldef\tempurl%
\url{https://doi.org/10.1037//0022-3514.66.2.398}
\showDOI{\tempurl}


\bibitem[Grail et~al\mbox{.}(2021)]%
        {grail2021globalizing}
\bibfield{author}{\bibinfo{person}{Quentin Grail}, \bibinfo{person}{Julien
  Perez}, {and} \bibinfo{person}{Eric Gaussier}.}
  \bibinfo{year}{2021}\natexlab{}.
\newblock \showarticletitle{Globalizing BERT-based transformer architectures
  for long document summarization}. In \bibinfo{booktitle}{\emph{Proceedings of
  the 16th Conference of the European Chapter of the Association for
  Computational Linguistics: Main Volume}}. \bibinfo{pages}{1792--1810}.
\newblock


\bibitem[Gupta et~al\mbox{.}(2022)]%
        {gupta2022automated}
\bibfield{author}{\bibinfo{person}{Anushka Gupta}, \bibinfo{person}{Diksha
  Chugh}, \bibinfo{person}{Rahul Katarya}, {et~al\mbox{.}}}
  \bibinfo{year}{2022}\natexlab{}.
\newblock \showarticletitle{Automated news summarization using transformers}.
\newblock In \bibinfo{booktitle}{\emph{Sustainable Advanced Computing}}.
  \bibinfo{publisher}{Springer}, \bibinfo{pages}{249--259}.
\newblock


\bibitem[Hermida and Young(2017)]%
        {hermida2017finding}
\bibfield{author}{\bibinfo{person}{Alfred Hermida} {and}
  \bibinfo{person}{Mary~Lynn Young}.} \bibinfo{year}{2017}\natexlab{}.
\newblock \showarticletitle{Finding the data unicorn: A hierarchy of hybridity
  in data and computational journalism}.
\newblock \bibinfo{journal}{\emph{Digital Journalism}} \bibinfo{volume}{5},
  \bibinfo{number}{2} (\bibinfo{year}{2017}), \bibinfo{pages}{159--176}.
\newblock


\bibitem[Isoaho et~al\mbox{.}(2021)]%
        {isoaho2021topic}
\bibfield{author}{\bibinfo{person}{Karoliina Isoaho}, \bibinfo{person}{Daria
  Gritsenko}, {and} \bibinfo{person}{Eetu M{\"a}kel{\"a}}.}
  \bibinfo{year}{2021}\natexlab{}.
\newblock \showarticletitle{Topic modeling and text analysis for qualitative
  policy research}.
\newblock \bibinfo{journal}{\emph{Policy Studies Journal}}
  \bibinfo{volume}{49}, \bibinfo{number}{1} (\bibinfo{year}{2021}),
  \bibinfo{pages}{300--324}.
\newblock


\bibitem[Jacobi et~al\mbox{.}(2016)]%
        {jacobi2016quantitative}
\bibfield{author}{\bibinfo{person}{Carina Jacobi}, \bibinfo{person}{Wouter
  Van~Atteveldt}, {and} \bibinfo{person}{Kasper Welbers}.}
  \bibinfo{year}{2016}\natexlab{}.
\newblock \showarticletitle{Quantitative analysis of large amounts of
  journalistic texts using topic modelling}.
\newblock \bibinfo{journal}{\emph{Digital journalism}} \bibinfo{volume}{4},
  \bibinfo{number}{1} (\bibinfo{year}{2016}), \bibinfo{pages}{89--106}.
\newblock


\bibitem[Jiang et~al\mbox{.}(2020)]%
        {jiang2020can}
\bibfield{author}{\bibinfo{person}{Zhengbao Jiang}, \bibinfo{person}{Frank~F
  Xu}, \bibinfo{person}{Jun Araki}, {and} \bibinfo{person}{Graham Neubig}.}
  \bibinfo{year}{2020}\natexlab{}.
\newblock \showarticletitle{How can we know what language models know?}
\newblock \bibinfo{journal}{\emph{Transactions of the Association for
  Computational Linguistics}}  \bibinfo{volume}{8} (\bibinfo{year}{2020}),
  \bibinfo{pages}{423--438}.
\newblock


\bibitem[Khanehzar et~al\mbox{.}(2019)]%
        {Khanehzar2019ModelingPF}
\bibfield{author}{\bibinfo{person}{Shima Khanehzar}, \bibinfo{person}{Andrew
  Turpin}, {and} \bibinfo{person}{Gosia Mikołajczak}.}
  \bibinfo{year}{2019}\natexlab{}.
\newblock \showarticletitle{Modeling Political Framing Across Policy Issues and
  Contexts}. In \bibinfo{booktitle}{\emph{ALTA}}.
\newblock


\bibitem[Kim and Wanta(2018)]%
        {kim2018news}
\bibfield{author}{\bibinfo{person}{Jeesun Kim} {and} \bibinfo{person}{Wayne
  Wanta}.} \bibinfo{year}{2018}\natexlab{}.
\newblock \showarticletitle{News framing of the US immigration debate during
  election years: Focus on generic frames}.
\newblock \bibinfo{journal}{\emph{The Communication Review}}
  \bibinfo{volume}{21}, \bibinfo{number}{2} (\bibinfo{year}{2018}),
  \bibinfo{pages}{89--115}.
\newblock


\bibitem[Liang et~al\mbox{.}(2022)]%
        {holistic}
\bibfield{author}{\bibinfo{person}{Percy Liang}, \bibinfo{person}{Rishi
  Bommasani}, \bibinfo{person}{Tony Lee}, \bibinfo{person}{Dimitris Tsipras},
  \bibinfo{person}{Dilara Soylu}, \bibinfo{person}{Michihiro Yasunaga},
  \bibinfo{person}{Yian Zhang}, \bibinfo{person}{Deepak Narayanan},
  \bibinfo{person}{Yuhuai Wu}, \bibinfo{person}{Ananya Kumar}, {et~al\mbox{.}}}
  \bibinfo{year}{2022}\natexlab{}.
\newblock \showarticletitle{Holistic evaluation of language models}.
\newblock \bibinfo{journal}{\emph{arXiv preprint arXiv:2211.09110}}
  (\bibinfo{year}{2022}).
\newblock


\bibitem[Liu et~al\mbox{.}(2021)]%
        {liu2021pre}
\bibfield{author}{\bibinfo{person}{Pengfei Liu}, \bibinfo{person}{Weizhe Yuan},
  \bibinfo{person}{Jinlan Fu}, \bibinfo{person}{Zhengbao Jiang},
  \bibinfo{person}{Hiroaki Hayashi}, {and} \bibinfo{person}{Graham Neubig}.}
  \bibinfo{year}{2021}\natexlab{}.
\newblock \showarticletitle{Pre-train, prompt, and predict: A systematic survey
  of prompting methods in natural language processing}.
\newblock  (\bibinfo{year}{2021}).
\newblock
\urldef\tempurl%
\url{https://doi.org/10.48550/ARXIV.2107.13586}
\showDOI{\tempurl}


\bibitem[Liu et~al\mbox{.}(2019)]%
        {liu2019detecting}
\bibfield{author}{\bibinfo{person}{Siyi Liu}, \bibinfo{person}{Lei Guo},
  \bibinfo{person}{Kate Mays}, \bibinfo{person}{Margrit Betke}, {and}
  \bibinfo{person}{Derry~Tanti Wijaya}.} \bibinfo{year}{2019}\natexlab{}.
\newblock \showarticletitle{Detecting frames in news headlines and its
  application to analyzing news framing trends surrounding US gun violence}. In
  \bibinfo{booktitle}{\emph{Proceedings of the 23rd conference on computational
  natural language learning (CoNLL)}}.
\newblock


\bibitem[Matthes and Kohring(2008)]%
        {deductive}
\bibfield{author}{\bibinfo{person}{Jörg Matthes} {and}
  \bibinfo{person}{Matthias Kohring}.} \bibinfo{year}{2008}\natexlab{}.
\newblock \showarticletitle{The Content Analysis of Media Frames: Toward
  Improving Reliability and Validity}.
\newblock \bibinfo{journal}{\emph{Journal of Communication}}
  \bibinfo{volume}{58} (\bibinfo{date}{06} \bibinfo{year}{2008}).
\newblock
\urldef\tempurl%
\url{https://doi.org/10.1111/j.1460-2466.2008.00384.x}
\showDOI{\tempurl}


\bibitem[Meyer et~al\mbox{.}(2022)]%
        {meyer2022we}
\bibfield{author}{\bibinfo{person}{Selina Meyer}, \bibinfo{person}{David
  Elsweiler}, \bibinfo{person}{Bernd Ludwig}, \bibinfo{person}{Marcos
  Fernandez-Pichel}, {and} \bibinfo{person}{David~E Losada}.}
  \bibinfo{year}{2022}\natexlab{}.
\newblock \showarticletitle{Do We Still Need Human Assessors? Prompt-Based
  GPT-3 User Simulation in Conversational AI}. In
  \bibinfo{booktitle}{\emph{Proceedings of the 4th Conference on Conversational
  User Interfaces}}. \bibinfo{pages}{1--6}.
\newblock


\bibitem[Middleton et~al\mbox{.}(2018)]%
        {middleton2018social}
\bibfield{author}{\bibinfo{person}{Stuart~E Middleton}, \bibinfo{person}{Symeon
  Papadopoulos}, {and} \bibinfo{person}{Yiannis Kompatsiaris}.}
  \bibinfo{year}{2018}\natexlab{}.
\newblock \showarticletitle{Social computing for verifying social media content
  in breaking news}.
\newblock \bibinfo{journal}{\emph{IEEE Internet Computing}}
  \bibinfo{volume}{22}, \bibinfo{number}{2} (\bibinfo{year}{2018}),
  \bibinfo{pages}{83--89}.
\newblock


\bibitem[Milosavljevi{\'c} and Vobi{\v{c}}(2021)]%
        {milosavljevic2021our}
\bibfield{author}{\bibinfo{person}{Marko Milosavljevi{\'c}} {and}
  \bibinfo{person}{Igor Vobi{\v{c}}}.} \bibinfo{year}{2021}\natexlab{}.
\newblock \showarticletitle{‘Our task is to demystify fears’: Analysing
  newsroom management of automation in journalism}.
\newblock \bibinfo{journal}{\emph{Journalism}} \bibinfo{volume}{22},
  \bibinfo{number}{9} (\bibinfo{year}{2021}), \bibinfo{pages}{2203--2221}.
\newblock


\bibitem[Monarch(2021)]%
        {munro2021human}
\bibfield{author}{\bibinfo{person}{R. Monarch}.}
  \bibinfo{year}{2021}\natexlab{}.
\newblock \bibinfo{booktitle}{\emph{Human-in-the-Loop Machine Learning: Active
  Learning and Annotation for Human-centered AI}}.
\newblock \bibinfo{publisher}{Manning}.
\newblock
\showISBNx{9781617296741}
\urldef\tempurl%
\url{https://books.google.ch/books?id=LCh0zQEACAAJ}
\showURL{%
\tempurl}


\bibitem[Nicholls and Culpepper(2021)]%
        {nicholls2021computational}
\bibfield{author}{\bibinfo{person}{Tom Nicholls} {and}
  \bibinfo{person}{Pepper~D Culpepper}.} \bibinfo{year}{2021}\natexlab{}.
\newblock \showarticletitle{Computational identification of media frames:
  Strengths, weaknesses, and opportunities}.
\newblock \bibinfo{journal}{\emph{Political Communication}}
  \bibinfo{volume}{38}, \bibinfo{number}{1-2} (\bibinfo{year}{2021}),
  \bibinfo{pages}{159--181}.
\newblock


\bibitem[Pan and Kosicki(1993)]%
        {pan1993framing}
\bibfield{author}{\bibinfo{person}{Zhongdang Pan} {and}
  \bibinfo{person}{Gerald~M Kosicki}.} \bibinfo{year}{1993}\natexlab{}.
\newblock \showarticletitle{Framing analysis: An approach to news discourse}.
\newblock \bibinfo{journal}{\emph{Political communication}}
  \bibinfo{volume}{10}, \bibinfo{number}{1} (\bibinfo{year}{1993}),
  \bibinfo{pages}{55--75}.
\newblock


\bibitem[Puri and Catanzaro(2019)]%
        {puri2019zero}
\bibfield{author}{\bibinfo{person}{Raul Puri} {and} \bibinfo{person}{Bryan
  Catanzaro}.} \bibinfo{year}{2019}\natexlab{}.
\newblock \showarticletitle{Zero-shot text classification with generative
  language models}.
\newblock \bibinfo{journal}{\emph{arXiv preprint arXiv:1912.10165}}
  (\bibinfo{year}{2019}).
\newblock


\bibitem[Qin and Eisner(2021)]%
        {qin2021learning}
\bibfield{author}{\bibinfo{person}{Guanghui Qin} {and} \bibinfo{person}{Jason
  Eisner}.} \bibinfo{year}{2021}\natexlab{}.
\newblock \showarticletitle{Learning how to ask: Querying lms with mixtures of
  soft prompts}.
\newblock \bibinfo{journal}{\emph{arXiv preprint arXiv:2104.06599}}
  (\bibinfo{year}{2021}).
\newblock


\bibitem[{Rabindra Lamsal}(2021)]%
        {bertsent}
\bibfield{author}{\bibinfo{person}{{Rabindra Lamsal}}.}
  \bibinfo{year}{2021}\natexlab{}.
\newblock \bibinfo{title}{{Sentiment Analysis of English Tweets with
  BERTsent}}.
\newblock
  \bibinfo{howpublished}{\url{https://huggingface.co/rabindralamsal/finetuned-bertweet-sentiment-analysis}}.
\newblock


\bibitem[Radford et~al\mbox{.}(2019)]%
        {radford2019language}
\bibfield{author}{\bibinfo{person}{Alec Radford}, \bibinfo{person}{Jeffrey Wu},
  \bibinfo{person}{Rewon Child}, \bibinfo{person}{David Luan},
  \bibinfo{person}{Dario Amodei}, \bibinfo{person}{Ilya Sutskever},
  {et~al\mbox{.}}} \bibinfo{year}{2019}\natexlab{}.
\newblock \showarticletitle{Language models are unsupervised multitask
  learners}.
\newblock \bibinfo{journal}{\emph{OpenAI blog}} \bibinfo{volume}{1},
  \bibinfo{number}{8} (\bibinfo{year}{2019}), \bibinfo{pages}{9}.
\newblock


\bibitem[Rai et~al\mbox{.}(2022)]%
        {RAI202298}
\bibfield{author}{\bibinfo{person}{Nishant Rai}, \bibinfo{person}{Deepika
  Kumar}, \bibinfo{person}{Naman Kaushik}, \bibinfo{person}{Chandan Raj}, {and}
  \bibinfo{person}{Ahad Ali}.} \bibinfo{year}{2022}\natexlab{}.
\newblock \showarticletitle{Fake News Classification using transformer based
  enhanced LSTM and BERT}.
\newblock \bibinfo{journal}{\emph{International Journal of Cognitive Computing
  in Engineering}}  \bibinfo{volume}{3} (\bibinfo{year}{2022}),
  \bibinfo{pages}{98--105}.
\newblock
\showISSN{2666-3074}
\urldef\tempurl%
\url{https://doi.org/10.1016/j.ijcce.2022.03.003}
\showDOI{\tempurl}


\bibitem[Rodelo(2021)]%
        {rodelo2021framing}
\bibfield{author}{\bibinfo{person}{Frida~V Rodelo}.}
  \bibinfo{year}{2021}\natexlab{}.
\newblock \showarticletitle{Framing of the Covid-19 pandemic and its
  organizational predictors}.
\newblock \bibinfo{journal}{\emph{Cuadernos. info}} \bibinfo{number}{50}
  (\bibinfo{year}{2021}), \bibinfo{pages}{91--112}.
\newblock


\bibitem[Scao et~al\mbox{.}(2022)]%
        {scao2022bloom}
\bibfield{author}{\bibinfo{person}{Teven~Le Scao}, \bibinfo{person}{Angela
  Fan}, \bibinfo{person}{Christopher Akiki}, \bibinfo{person}{Ellie Pavlick},
  \bibinfo{person}{Suzana Ili{\'c}}, \bibinfo{person}{Daniel Hesslow},
  \bibinfo{person}{Roman Castagn{\'e}}, \bibinfo{person}{Alexandra~Sasha
  Luccioni}, \bibinfo{person}{Fran{\c{c}}ois Yvon}, \bibinfo{person}{Matthias
  Gall{\'e}}, {et~al\mbox{.}}} \bibinfo{year}{2022}\natexlab{}.
\newblock \showarticletitle{Bloom: A 176b-parameter open-access multilingual
  language model}.
\newblock \bibinfo{journal}{\emph{arXiv preprint arXiv:2211.05100}}
  (\bibinfo{year}{2022}).
\newblock


\bibitem[Semetko and Valkenburg(2000)]%
        {semetko}
\bibfield{author}{\bibinfo{person}{Holli Semetko} {and} \bibinfo{person}{Patti
  Valkenburg}.} \bibinfo{year}{2000}\natexlab{}.
\newblock \showarticletitle{Framing European Politics: A Content Analysis of
  Press and Television News}.
\newblock \bibinfo{journal}{\emph{Journal of Communication}}
  \bibinfo{volume}{50} (\bibinfo{date}{06} \bibinfo{year}{2000}),
  \bibinfo{pages}{93 -- 109}.
\newblock
\urldef\tempurl%
\url{https://doi.org/10.1111/j.1460-2466.2000.tb02843.x}
\showDOI{\tempurl}


\bibitem[Shin et~al\mbox{.}(2021)]%
        {shin2021constrained}
\bibfield{author}{\bibinfo{person}{Richard Shin},
  \bibinfo{person}{Christopher~H Lin}, \bibinfo{person}{Sam Thomson},
  \bibinfo{person}{Charles Chen}, \bibinfo{person}{Subhro Roy},
  \bibinfo{person}{Emmanouil~Antonios Platanios}, \bibinfo{person}{Adam Pauls},
  \bibinfo{person}{Dan Klein}, \bibinfo{person}{Jason Eisner}, {and}
  \bibinfo{person}{Benjamin Van~Durme}.} \bibinfo{year}{2021}\natexlab{}.
\newblock \showarticletitle{Constrained language models yield few-shot semantic
  parsers}.
\newblock \bibinfo{journal}{\emph{arXiv preprint arXiv:2104.08768}}
  (\bibinfo{year}{2021}).
\newblock


\bibitem[Sidiropoulos and Veglis(2017)]%
        {articlecscw}
\bibfield{author}{\bibinfo{person}{Efstathios Sidiropoulos} {and}
  \bibinfo{person}{Andreas Veglis}.} \bibinfo{year}{2017}\natexlab{}.
\newblock \showarticletitle{Computer Supported Collaborative Work trends on
  Media Organizations: Mixing Qualitative and Quantitative Approaches}.
\newblock \bibinfo{journal}{\emph{Studies in Media and Communication}}
  \bibinfo{volume}{5} (\bibinfo{date}{04} \bibinfo{year}{2017}),
  \bibinfo{pages}{63}.
\newblock
\urldef\tempurl%
\url{https://doi.org/10.11114/smc.v5i1.2279}
\showDOI{\tempurl}


\bibitem[Strubell et~al\mbox{.}(2019)]%
        {strubell2019energy}
\bibfield{author}{\bibinfo{person}{Emma Strubell}, \bibinfo{person}{Ananya
  Ganesh}, {and} \bibinfo{person}{Andrew McCallum}.}
  \bibinfo{year}{2019}\natexlab{}.
\newblock \showarticletitle{Energy and policy considerations for deep learning
  in NLP}.
\newblock \bibinfo{journal}{\emph{arXiv preprint arXiv:1906.02243}}
  (\bibinfo{year}{2019}).
\newblock


\bibitem[Tamkin et~al\mbox{.}(2021)]%
        {tamkin2021understanding}
\bibfield{author}{\bibinfo{person}{Alex Tamkin}, \bibinfo{person}{Miles
  Brundage}, \bibinfo{person}{Jack Clark}, {and} \bibinfo{person}{Deep
  Ganguli}.} \bibinfo{year}{2021}\natexlab{}.
\newblock \showarticletitle{Understanding the capabilities, limitations, and
  societal impact of large language models}.
\newblock \bibinfo{journal}{\emph{arXiv preprint arXiv:2102.02503}}
  (\bibinfo{year}{2021}).
\newblock


\bibitem[Trinh and Le(2018)]%
        {trinh2018simple}
\bibfield{author}{\bibinfo{person}{Trieu~H Trinh} {and} \bibinfo{person}{Quoc~V
  Le}.} \bibinfo{year}{2018}\natexlab{}.
\newblock \showarticletitle{A simple method for commonsense reasoning}.
\newblock \bibinfo{journal}{\emph{arXiv preprint arXiv:1806.02847}}
  (\bibinfo{year}{2018}).
\newblock


\bibitem[Tsimpoukelli et~al\mbox{.}(2021)]%
        {tsimpoukelli2021multimodal}
\bibfield{author}{\bibinfo{person}{Maria Tsimpoukelli},
  \bibinfo{person}{Jacob~L Menick}, \bibinfo{person}{Serkan Cabi},
  \bibinfo{person}{SM Eslami}, \bibinfo{person}{Oriol Vinyals}, {and}
  \bibinfo{person}{Felix Hill}.} \bibinfo{year}{2021}\natexlab{}.
\newblock \showarticletitle{Multimodal few-shot learning with frozen language
  models}.
\newblock \bibinfo{journal}{\emph{Advances in Neural Information Processing
  Systems}}  \bibinfo{volume}{34} (\bibinfo{year}{2021}),
  \bibinfo{pages}{200--212}.
\newblock


\bibitem[Vannoy and Palvia(2010)]%
        {vannoy2010social}
\bibfield{author}{\bibinfo{person}{Sandra~A Vannoy} {and}
  \bibinfo{person}{Prashant Palvia}.} \bibinfo{year}{2010}\natexlab{}.
\newblock \showarticletitle{The social influence model of technology adoption}.
\newblock \bibinfo{journal}{\emph{Commun. ACM}} \bibinfo{volume}{53},
  \bibinfo{number}{6} (\bibinfo{year}{2010}), \bibinfo{pages}{149--153}.
\newblock


\bibitem[Yl{\"a}-Anttila et~al\mbox{.}(2022)]%
        {yla2022topic}
\bibfield{author}{\bibinfo{person}{Tuukka Yl{\"a}-Anttila},
  \bibinfo{person}{Veikko Eranti}, {and} \bibinfo{person}{Anna Kukkonen}.}
  \bibinfo{year}{2022}\natexlab{}.
\newblock \showarticletitle{Topic modeling for frame analysis: A study of media
  debates on climate change in India and USA}.
\newblock \bibinfo{journal}{\emph{Global Media and Communication}}
  \bibinfo{volume}{18}, \bibinfo{number}{1} (\bibinfo{year}{2022}),
  \bibinfo{pages}{91--112}.
\newblock


\end{thebibliography}










\end{document}